\documentclass[runningheads]{llncs}
\usepackage[T1]{fontenc}
\usepackage{graphicx,verbatim,multirow}
\usepackage{amsmath, amssymb}
\usepackage{xcolor}
\usepackage{array}
\begin{document}
%
\title{PromptDx: Differentiable Prompt Tuning for Multimodal In-Context Alzheimer's Diagnosis}
%

\author{%
  Lujia Zhong\textsuperscript{1,2}, Yihao Xia\textsuperscript{1}, Shuo Huang\textsuperscript{1,3},\\
  Jianwei Zhang\textsuperscript{1,2}, Yonggang Shi\textsuperscript{1,2,3}
  and for the Alzheimer’s Disease Neuroimaging Initiative
}
\institute{
\textsuperscript{1} Stevens Neuroimaging and Informatics Institute, Keck School of Medicine, University of Southern California \\
  \textsuperscript{2} Ming Hsieh Department of Electrical and Computer Engineering, Viterbi School of
Engineering, University of Southern California \\
  \textsuperscript{3} Alfred E. Mann Department of Biomedical Engineering, Viterbi School of
Engineering, University of Southern California \\}

\maketitle              
\begin{abstract}
Deep learning models in medical imaging typically operate as parametric memory, diagnosing patients by recalling fixed knowledge learned during training. This contrasts sharply with clinical practice, where physicians employ analogical reasoning to diagnose new cases by referencing similar records from past exemplars. While In-Context Learning (ICL) frameworks such as Tabular Prior-Fitted Networks (TabPFN) offer a promising diagnosis-by-reference paradigm, they are designed with tabular-specific inductive priors and rely on non-differentiable preprocessing pipelines, leading to manifold mismatch and gradient fracture when applied to heterogeneous multimodal data. To address these limitations, we propose PromptDx, a novel diagnosis-by-reference framework that leverages a pre-trained TabPFN as an ICL engine while enabling seamless integration with multimodal representations. Our core contribution is a \textbf{D}ifferentiable \textbf{P}rompt \textbf{T}uning (DPT) mechanism that aligns a Masked Multimodal Modeling module with the pre-trained ICL engine. By training a lightweight adapter as a differentiable surrogate for the engine's non-differentiable preprocessors, we enable an end-to-end optimization of multimodal prompts within the ICL paradigm. We validate our method on the Alzheimer's Disease Neuroimaging Initiative (ADNI) dataset using 3D MRI and tabular biomarkers. Experiments demonstrate that our approach outperforms traditional parametric baselines. Notably, our method achieves superior performance using only 1\% context samples compared to 30\% in standard ICL, demonstrating exceptional manifold condensation ability. We further validate the generalizability of our DPT framework across six tabular datasets with diverse scales. Overall, our method offers a more data-efficient and clinically aligned paradigm for Alzheimer's Disease diagnosis.

\end{abstract}
\section{Introduction}

When a clinician evaluates a patient for Alzheimer’s Disease (AD), they do not rely solely on a static set of memorized rules. Instead, they employ \textit{analogical reasoning}: comparing the patient's symptoms and phenotypes against a contextual memory of prior cases derived from years of experience. Such analogical reasoning is a hallmark of Case-Based Reasoning (CBR) \cite{aamodt1994case}. This diagnosis-by-reference allows physicians to make robust inferences even with limited data, dynamically retrieving relevant exemplars to ground their decision-making.

In contrast, the dominant paradigm in medical Artificial Intelligence (AI) is \textit{parametric learning}. Deep learning models, such as Convolutional Neural Networks \cite{he2016deep,huang2017densely,hu2018squeeze}, Vision Transformers \cite{dosovitskiy2020image,he2023swinunetr,swamy2024intrinsic,fedus2022switch,zhang2022mmformer,tsai2019multimodal}, and Mixture Models \cite{jin2024moe++,yun2024flex,xin2025i2moe} compress training data into fixed neural weights. They require large amounts of data for training, and once deployed, these models function as parametric memory; they cannot explicitly reference training examples or adapt to new distributions without retraining or finetuning. Consequently, they often struggle with generalizability and the data scarcity inherent in medical applications, especially AD analysis.

\textit{In-Context Learning} (ICL) offers a compelling alternative. Pioneered by Large Language Models (LLMs) \cite{brown2020language} and formalized for tabular data by Prior-Fitted Networks (PFNs) like TabPFN \cite{hollmann2025accurate}, ICL allows a model to make predictions by conditioning on a support set of examples at inference time without updating model weights. However, applying ICL to high-dimensional multi-modal neuroimaging data presents a critical challenge: a \textit{manifold mismatch}. Powerful ICL engines like TabPFN are typically designed for tabular priors and cannot directly digest visual images or volumes. Furthermore, the non-differentiable preprocessing pipelines inherent in ICL engines create a gradient fracture, preventing an end-to-end optimization to generate compatible prompts.

In this work, we propose a novel framework along with a differentiable prompt tuning approach that shifts the learning objective from training a disease classifier to \textit{learning to prompt}. 
Specifically, we first employ a masked multimodal modeling module to extract fused representations from 3D MRI and tabular data. Second, we employ an adapter trained with dual objectives, aligning the multimodal features with the ICL engine's expected manifold while optimizing them as soft prompts to improve AD diagnosis. Our contributions are threefold:
\begin{enumerate}
    \item We introduce the first diagnosis-by-reference framework for multimodal AD diagnosis and formulate it as an ICL problem.
    \item We propose a differentiable prompt tuning solution to ICL manifold mismatch and gradient fracture, treating features as soft prompts that can be optimized to improve classification performance.
    \item We demonstrate our approach on the ADNI dataset and 6 diverse tabular datasets, showing that our method outperforms baselines in accuracy and data efficiency while offering generality beyond AD diagnosis.
\end{enumerate}

\section{Method}

\begin{figure}[t]
\centering
\includegraphics[width=1.0\textwidth]{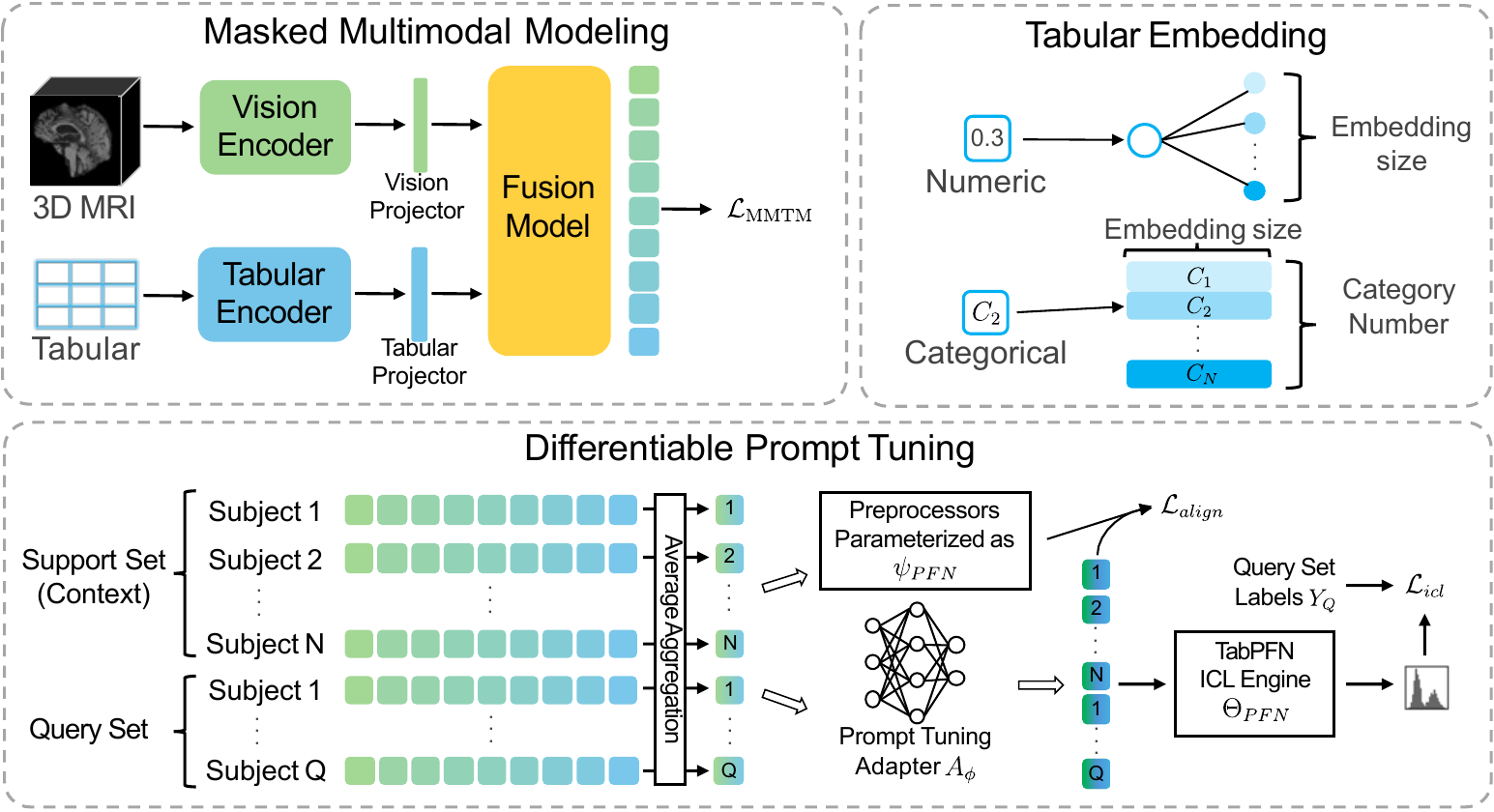}
\caption{Overview of our proposed PromptDx framework. A multimodal representation learning stage to acquire fused multimodal features from 3D MRI and tabular data, where specialized tabular embedding is built within the tabular encoder for structured data. The proposed DPT is applied for soft prompt tuning on fused features.}
\label{fig:workflow}
\end{figure}

Our framework unifies robust multimodal representation learning with the flexibility of ICL. As shown in Fig. \ref{fig:workflow}, the pipeline proceeds in two stages: (1) \textbf{M}asked \textbf{M}ul\textbf{T}imodal \textbf{M}odeling (MMTM) pre-training to fuse 3D MRI and tabular data into a coherent feature space, and (2) \textbf{D}ifferentiable \textbf{P}rompt \textbf{T}uning (DPT) to align these features with a pre-trained ICL inference engine.

\subsection{Problem Formulation}
Standard supervised learning aims to optimize a parametric function $f_{\theta}: \mathcal{X} \rightarrow \mathcal{Y}$ by compressing diagnostic knowledge into static model weights, where $\mathcal{X}$ denotes the input set of multimodal data, and $\mathcal{Y}$ represents the set of diagnostic labels. In contrast, we adopt a diagnosis-by-reference formulation, which treats the classification of a query patient $x_q$ as a posterior estimation problem conditioned on a support set of labeled exemplars. Let $\mathcal{S}$ and $\mathcal{Q}$ denote the index set of support and query samples, respectively, within an ICL episode. The support dataset is defined as $\mathcal{D}_S = \{(x_i, y_i)\}_{i \in \mathcal{S}}$, where each $x_i \in \mathcal{X}$ is a reference patient's profile and $y_i \in \mathcal{Y}$ is the corresponding ground-truth. $$\hat{y}_q = \arg\max_{y\in\mathcal{Y}} P\big(y \mid x_q, D_S\big)$$

Given the high-dimensional nature of heterogeneous multimodal inputs (3D MRI and tabular biomarkers), we employ the proposed MMTM module to extract fused representations $h = \text{MMTM}(x) \in \mathcal{H}$, which capture the cross-modal correlations between neuroanatomy and clinical phenotypes. Ideally, the fusion module could be optimized end-to-end using the ICL classification objective, as ICL engines can potentially provide superior relational guidance for feature extraction compared to traditional parametric heads (e.g., MLPs). However, two fundamental barriers exist: manifold mismatch and gradient fracture. 

First, powerful ICL engines (e.g., TabPFN) are typically pre-trained on specific tabular priors and expect inputs within a restricted manifold space $\mathcal{Z}_{ICL}$, which does not naturally align with the feature space $\mathcal{H}$ of the MMTM. This is usually handled by TabPFN's preprocessing pipelines $\psi_{PFN}$. However, the $\psi_{PFN}$ used by ICL engines is non-differentiable, relying on sorting and ranking for quantile transformations. This non-differentiability breaks the gradient flow and prevents ICL engines from directly guiding the MMTM via backpropagation.

To bridge this gap, we propose the DPT framework. We introduce a learnable adapter $A_{\phi}: \mathcal{H} \rightarrow \mathcal{Z}_{ICL}$ that functions as a differentiable surrogate for the non-differentiable pipeline $\psi_{PFN}$. The adapter transforms fused multimodal features $h$ into soft latent prompts $z = A_{\phi}(h)$. This formulation allows the query and support set to be projected into an ICL-compatible manifold as $z_q = A_{\phi}(h_q)$ and $\mathcal{Z}_S = \{A_{\phi}(h_i)\}_{i \in \mathcal{S}}$. The corresponding support labels are defined as $\mathcal{Y}_S = \{y_i\}_{i \in \mathcal{S}}$. Consequently, the diagnostic inference is performed as: $$\hat{y} = \arg\max_y P(y | z_q, \mathcal{Z}_S, \mathcal{Y}_S; \Theta_{PFN})$$ where $\Theta_{PFN}$ represents the pre-trained parameters of the ICL engine. By establishing a continuous gradient path from the decision layer back to the MMTM, we can theoretically optimize the MMTM module and $A_{\phi}$ to ensure that extracted features are both statistically compatible with the engine's prior and semantically discriminative for Alzheimer's diagnosis.

In practice, full end-to-end updates of the MMTM module within each ICL episode are computationally prohibitive due to GPU memory constraints and the need for repeated inference on the entire support set. Therefore, we adopt a decoupled two-stage strategy: (1) We train the MMTM module via self-supervised mask modeling to acquire robust representations, and (2) we focus the ICL-guided end-to-end optimization on the adapter $A_{\phi}$. This modular approach retains the advantages of ICL-guided feature refinement through soft prompt tuning while ensuring training efficiency.

\subsection{Masked Multimodal Modeling}

The representation learning stage is designed to extract structural biomarkers from 3D MRI and integrate them with diverse clinical phenotypes. Given a 3D MRI $X_{img} \in \mathbb{R}^{D \times H \times W}$, we first partition it into non-overlapping 3D square patches of size 16, which are then projected into a sequence of visual embeddings $E_v$ \cite{dosovitskiy2020image}. Simultaneously, as shown in Fig. \ref{fig:workflow}, clinical tabular data $X_{tab} \in \mathbb{R}^d$ is transformed into a sequence of embeddings $E_t$ through learnable embedding layers for categorical variables and MLPs for continuous ones. Essentially, visual and tabular embeddings are processed by 4-layer and 2-layer Transformer encoders, respectively, to capture intra-modal dependencies before fusion.

These embeddings are then concatenated into a unified sequence $Z_0 = [E_v; E_t]$ and fed into a 24-layer Transformer fusion module, where the bi-directional attention mechanism allows the model to capture complex, non-linear cross-modal correlations. To establish a robust joint distribution $P(X_{img}, X_{tab})$ between neuroanatomy and phenotype, we randomly mask a subset of visual and clinical input embeddings during training with different ratios of $\rho_{img}$ and $\rho_{tab}$, requiring the model to reconstruct the masked inputs based on the multimodal context \cite{he2022masked}. Specifically, the fused embeddings are processed by modality-specific decoders, comprising an MLP layer for vision and embedding layers/MLPs for table, to map the features back to their original input dimensions. The final reconstruction loss $\mathcal{L}_{\text{MMTM}}$ comprises two L2 losses for visual and tabular data, respectively. 

Upon completion, the fused representations from the final layer are averaged channel-wise to form a high-density multimodal feature vector, ensuring that the ICL inference engine receives a feature set already rich in cross-modal context.

\subsection{Differentiable Prompt Tuning}

In this stage, we treat the pre-trained MMTM as a feature extractor and the TabPFN as an ICL classifier. We employ a composite objective function that ensures the generated soft prompts \cite{lester2021power,liu2024gpt} are compatible with the TabPFN's ICL engine $\Theta_{PFN}$ and also optimally optimized for AD diagnosis.

\textit{Manifold Alignment Loss}: This term forces the adapter $A_\phi$, parameterized by 6-layer MLPs, to function as a differentiable surrogate for the non-differentiable pipeline $\psi_{PFN}$ and keeps the adapter's output to remain within the specific manifold that the ICL engine was pre-trained to interpret.

\begin{equation*}
\mathcal{L}_{align} = \frac{1}{|\mathcal{T}|} \sum_{i \in \mathcal{T}}
\left\| A_\phi(h_i) - \psi_{PFN}(h_i) \right\|_1,
\quad \text{where } \mathcal{T} = \mathcal{S} \cup \mathcal{Q}.
\end{equation*}

\textit{ICL Classification Loss}: With the adapter providing a valid surrogate transformation, we optimize the adapter for classification performance. The adapter transforms the base representations into soft latent prompts $\mathcal{Z}_{\mathcal{S}}$ for the support set and $z_j = A_\phi(h_j)$ for each query $j \in \mathcal{Q}$. These modified prompts are fed into the TabPFN model $\Theta_{PFN}$ to estimate the posterior.

\begin{equation*}
\mathcal{L}_{icl} = -\frac{1}{|\mathcal{Q}|} \sum_{j \in \mathcal{Q}} \sum_{c=1}^C \mathbb{I}\{y_j=c\} \log P(y_j=c | z_j, \mathcal{Z}_{S}, \mathcal{Y}_{S}; \Theta_{PFN})
\end{equation*}

where $C$ denotes the number of diagnostic classes; $\mathbb{I}\{y_j=c\}$ denotes the indicator function that equals 1 if the ground-truth label of sample $j$ is class $c$, and 0 otherwise; and $\mathcal{Y}_{S}$ represents the ground-truth labels of the support set. The final objective is formulated as $\mathcal{L}_{total} = \lambda_{icl} \mathcal{L}_{icl} + \lambda_{align} \mathcal{L}_{align}$, where $\lambda_{icl}$ and $\lambda_{align}$ are hyperparameters controlling the contribution of each term.

\begin{table}[t]
\centering
\caption{Performance comparison on the ADNI test set. The \textbf{best results} and the \underline{second best} are highlighted. Standard deviations are reported in subscript.}
\label{tab:main_results}
\resizebox{\textwidth}{!}{%
\begin{tabular}{l|cc|ccccc}
\hline
\textbf{Method} & \textbf{Img} & \textbf{Tab} & \textbf{Acc. (\%)} & \textbf{F1 (\%)} & \textbf{AUC (\%)} & \textbf{Sens. (\%)} & \textbf{Spec. (\%)} \\ \hline
\multicolumn{8}{l}{\textit{Unimodal Baselines}} \\ \hline
ResNet \cite{he2016deep} & \checkmark & & 46.91$_{\pm 3.90}$ & 39.27$_{\pm 3.17}$ & 69.70$_{\pm 2.17}$ & 46.55$_{\pm 3.22}$ & 71.96$_{\pm 2.37}$ \\
DenseNet \cite{huang2017densely} & \checkmark & & 46.06$_{\pm 4.89}$ & 39.43$_{\pm 4.33}$ & 68.55$_{\pm 2.22}$ & 45.39$_{\pm 3.85}$ & 72.20$_{\pm 3.54}$ \\
SE-ResNet \cite{hu2018squeeze} & \checkmark & & 46.91$_{\pm 4.78}$ & 43.27$_{\pm 3.51}$ & 62.86$_{\pm 1.04}$ & 44.34$_{\pm 0.50}$ & 71.31$_{\pm 0.94}$ \\
SwinUNETR \cite{he2023swinunetr} & \checkmark & & 48.53$_{\pm 1.65}$ & 42.32$_{\pm 1.34}$ & 64.77$_{\pm 0.96}$ & 43.29$_{\pm 2.85}$ & 71.89$_{\pm 0.51}$ \\
XGBoost \cite{chen2016xgboost} & & \checkmark & 57.68$_{\pm 0.26}$ & 54.51$_{\pm 0.39}$ & 74.83$_{\pm 0.23}$ & 53.90$_{\pm 0.36}$ & 77.40$_{\pm 0.09}$ \\
AutoGluon \cite{erickson2020autogluon} & & \checkmark & 60.49$_{\pm 1.52}$ & 58.01$_{\pm 0.70}$ & 77.77$_{\pm 0.18}$ & 57.96$_{\pm 1.05}$ & 79.01$_{\pm 0.62}$ \\
TabPFN-v2.5 \cite{hollmann2025accurate} & & \checkmark & 61.57$_{\pm 0.39}$ & \underline{59.19}$_{\pm 0.25}$ & 78.84$_{\pm 0.04}$ & 58.64$_{\pm 0.20}$ & 78.20$_{\pm 0.20}$ \\
\hline
\multicolumn{8}{l}{\textit{Multimodal Baselines}} \\ \hline
InterpretCC \cite{swamy2024intrinsic} & \checkmark & \checkmark & \underline{62.42}$_{\pm 0.76}$ & 53.29$_{\pm 2.02}$ & 75.51$_{\pm 0.26}$ & 57.24$_{\pm 1.71}$ & 78.11$_{\pm 0.68}$ \\
MoE++ \cite{jin2024moe++} & \checkmark & \checkmark & 60.25$_{\pm 3.37}$ & 58.45$_{\pm 2.62}$ & 78.64$_{\pm 1.56}$ & 59.10$_{\pm 3.05}$ & \underline{79.32}$_{\pm 1.63}$ \\
SwitchGate \cite{fedus2022switch} & \checkmark & \checkmark & 60.87$_{\pm 0.63}$ & 57.15$_{\pm 2.18}$ & \underline{78.87}$_{\pm 0.69}$ & 57.74$_{\pm 2.05}$ & 78.71$_{\pm 0.80}$ \\
mmFormer \cite{zhang2022mmformer} & \checkmark & \checkmark & 53.95$_{\pm 2.29}$ & 51.65$_{\pm 2.98}$ & 70.41$_{\pm 2.12}$ & 55.24$_{\pm 0.45}$ & 76.78$_{\pm 0.54}$ \\
MulT \cite{tsai2019multimodal} & \checkmark & \checkmark & 59.42$_{\pm 2.67}$ & 56.42$_{\pm 2.77}$ & 76.18$_{\pm 2.04}$ & 58.72$_{\pm 2.43}$ & 78.55$_{\pm 1.27}$ \\
Flex-MoE \cite{yun2024flex} & \checkmark & \checkmark & 61.05$_{\pm 1.04}$ & 58.65$_{\pm 0.44}$ & 75.27$_{\pm 1.21}$ & 58.58$_{\pm 1.30}$ & 78.56$_{\pm 0.58}$ \\
I2MoE \cite{xin2025i2moe} & \checkmark & \checkmark & 61.64$_{\pm 0.90}$ & 56.72$_{\pm 2.69}$ & 78.62$_{\pm 0.33}$ & \underline{59.49}$_{\pm 0.91}$ & 78.96$_{\pm 0.58}$ \\ \hline
\textbf{Ours} & \checkmark & \checkmark & \textbf{63.44}$_{\pm 0.82}$ & \textbf{62.63}$_{\pm 0.82}$ & \textbf{79.39}$_{\pm 0.38}$ & \textbf{62.88}$_{\pm 0.92}$ & \textbf{80.53}$_{\pm 0.45}$ \\ \hline
\end{tabular}%
}
\end{table}

\section{Experiments}

\subsection{Experimental Setting}

\begin{figure}[t]
\centering
\includegraphics[width=0.7\textwidth]{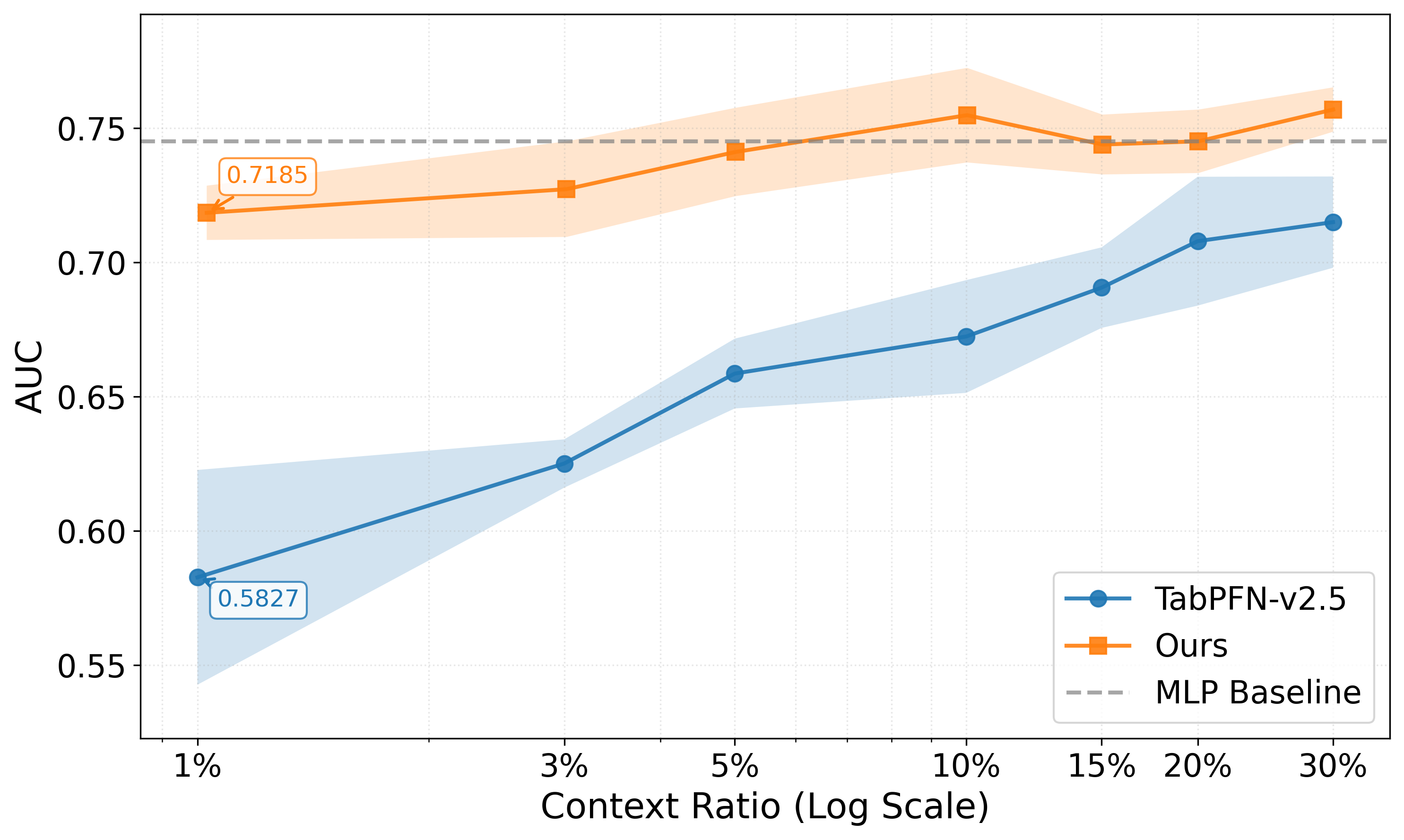}
\caption{Performance comparison on AUC with different context ratios.}
\label{fig:exp2_baseline_auc}
\end{figure}

We evaluate our framework on the Alzheimer’s Disease Neuroimaging Initiative (ADNI) dataset \cite{mueller2005alzheimer,jack2008alzheimer}, comprising 2240 subjects (1000 CN, 770 MCI, and 470 AD). The input modalities consist of 3D T1-weighted MRI (size $128 \times 128 \times 128$, skull-stripped and preprocessed via FreeSurfer \cite{fischl2012freesurfer}) and tabular data with 172 columns (medical history, neuro exam, demographics, vitals, APOE results, and CSF biomarkers), where the neuro exam does not include cognitive tests. We followed the same preprocessing procedure as described in Flex-MoE \cite{yun2024flex}. All subjects have MRI data, and missing tabular values were addressed using mean imputation. The dataset was randomly partitioned into training (70\%), validation (15\%), and testing (15\%) sets. We perform a three-class classification (CN vs. MCI vs. AD) and report performance using Accuracy, F1-score, AUC, Sensitivity, and Specificity. Results are reported as mean ± standard deviation across three independent runs to ensure reproducibility.

All experiments were conducted on an NVIDIA A6000 GPU using the Adam optimizer with default settings. We compare against various state-of-the-art unimodal and multimodal baselines, all implemented using the optimal hyperparameters reported in their original works for fair comparison. For the ADNI dataset, MMTM pre-training was performed for 100 epochs (lr = $2 \times 10^{-4}$, batch size = 32) with mask ratios of 0.05 (tabular) and 0.75 (MRI). In the DPT stage, we trained for 100 epochs (lr = $1 \times 10^{-3}$) with weight factors $\lambda_{align}=1$ and $\lambda_{icl}=0.01$. The support (context) set size for DPT training is 70\% of the training data, and the remaining 30\% is used for the query set; during inference, we use all training data as context. For the experiments in Table 2, we set both weight factors to be 1, and other settings remain the same unless specified otherwise.

\subsection{Results}

The evaluation of proposed PromptDx on the ADNI dataset demonstrates the superiority of our framework over existing paradigms. As detailed in Table \ref{tab:main_results}, our method achieves state-of-the-art performance with an Accuracy of 63.44\%, an F1-score of 62.63\%, and an AUC of 79.39\%. These results significantly outperform unimodal baselines such as SwinUNETR, as well as advanced multimodal models like Flex-MoE. This performance gain suggests that shifting from fixed parametric memory to a diagnosis-by-reference approach allows the model to better utilize analogical reasoning. By conditioning predictions on the context of exemplars, PromptDx captures complex cross-modal and cross-subject correlations that previous methods often struggle to parameterize.

A critical advantage of our proposed DPT mechanism is its exceptional manifold condensation and data efficiency. As illustrated in Fig. \ref{fig:exp2_baseline_auc}, our method with only 1\% context samples achieves an AUC exceeding the standard TabPFN utilizing 30\% samples. With only 10\% context samples, our method outperforms the MLP baseline trained on all data. This efficiency indicates that the DPT adapter effectively distills high-dimensional multimodal features into highly informative soft prompts optimized for the ICL engine's internal reasoning.

The ablation study presented in Table \ref{tab:ablation} highlights the necessity of each component in our framework. The removal of MMTM pre-training leads to a substantial decline in F1-score (from 62.63\% to 58.73\%), underscoring its role in establishing a robust joint distribution between anatomy and phenotype. Notably, excluding the Manifold Alignment Loss results in the most significant performance degradation, with the F1-score dropping to 57.42\%. This drop demonstrates that it is essential to overcome the manifold mismatch between the fused representations and the ICL engine input space. The other two components of the DPT adapter and ICL classification loss also contribute to our full method, demonstrating the power of ICL-guided soft prompt tuning. Overall, by integrating a differentiable surrogate, our framework ensures that multimodal representations are warped into a manifold where the relationships between query patients and clinical exemplars are most diagnostically distinguishable.

\begin{table}[t]
\centering
\caption{Ablation study on the ADNI dataset, showing the contribution of each key component within our framework.}
\label{tab:ablation}
\begin{tabular}{l|cc}
\hline
\textbf{Method} & \textbf{Acc. (\%)} & \textbf{F1 (\%)} \\ \hline
\textbf{Full Method (Ours)} & \textbf{63.44} & \textbf{62.63} \\
w/o MMTM Pre-training & 60.68 & 58.73 \\
w/o DPT Adapter & 61.82 & 60.62 \\
w/o Alignment Loss ($\mathcal{L}_{align}$) & 60.96 & 57.42 \\
w/o Classification Loss ($\mathcal{L}_{icl}$) & 61.25 & 59.79 \\
\hline
\end{tabular}
\end{table}

Beyond neuroimaging, the generalizability of this framework is validated across six diverse tabular datasets in Table \ref{tab:generalizability}, including CovType, Digits, and synthetic datasets with varying scales. Our method consistently outperforms both TabPFN-v2.5 and its finetuned version, confirming that our DPT strategy is a robust solution for diverse data distributions beyond AD diagnosis task.

\begin{table}[t]
\centering
\caption{Generalizability evaluation of the proposed DPT on diverse tabular datasets.}
\label{tab:generalizability}
\begin{tabular}{l|ccccc}
\hline
\textbf{Dataset} & \textbf{Acc. (\%)} & \textbf{F1 (\%)} & \textbf{AUC (\%)} & \textbf{Sens. (\%)} & \textbf{Spec. (\%)} \\ \hline
\multicolumn{6}{l}{\textit{CovType (3000 samples, 512 context, 54 features, 7 classes)}} \\ \hline
TabPFN-v2.5 & 70.83 & 47.50 & 92.73 & 47.31 & 93.15 \\
TabPFN-v2.5-finetune & 71.81 & 47.29 & \textbf{92.90} & 47.39 & 93.39 \\
\textbf{Ours (DPT)} & \textbf{73.33} & \textbf{51.47} & 91.22 & \textbf{50.50} & \textbf{93.78} \\ \hline
\multicolumn{6}{l}{\textit{Digits (1797 samples, 128 context, 64 features, 10 classes)}} \\ \hline
TabPFN-v2.5 & 90.56 & 90.29 & 99.21 & 90.49 & 98.95 \\
TabPFN-v2.5-finetune & 92.24 & 91.98 & 99.28 & 92.16 & 99.14 \\
\textbf{Ours (DPT)} & \textbf{94.17} & \textbf{94.04} & \textbf{99.70} & \textbf{94.12} & \textbf{99.35} \\ \hline
\multicolumn{6}{l}{\textit{Synthetic 1 (10000 samples, 2000 context, 100 features, 5 balanced classes)}} \\ \hline
TabPFN-v2.5 & 74.40 & 74.30 & 93.75 & 74.38 & 93.60 \\
TabPFN-v2.5-finetune & 76.65 & 76.60 & 94.14 & 76.63 & 94.16 \\
\textbf{Ours (DPT)} & \textbf{87.05} & \textbf{87.04} & \textbf{98.01} & \textbf{87.04} & \textbf{96.76} \\ \hline
\multicolumn{6}{l}{\textit{Synthetic 2 (50000 samples, 8000 context, 32 features, 10 balanced classes)}} \\ \hline
TabPFN-v2.5 & 83.63 & 83.60 & 98.02 & 83.63 & 98.18 \\
TabPFN-v2.5-finetune & 84.78 & 84.77 & 98.12 & 84.78 & 98.31 \\
\textbf{Ours (DPT)} & \textbf{88.32} & \textbf{88.30} & \textbf{98.71} & \textbf{88.32} & \textbf{98.70} \\ \hline
\end{tabular}%
\end{table}

\section{Conclusion}
In this work, we introduced PromptDx, a novel framework that shifts Alzheimer’s Disease diagnosis from static parametric memory to analogical diagnosis-by-reference using In-Context Learning. By proposing a Differentiable Prompt Tuning mechanism and a learnable surrogate adapter, we successfully bridged the manifold mismatch between high-dimensional multimodal data and pretrained ICL engines. Our experiments on the ADNI dataset demonstrate state-of-the-art performance and exceptional manifold condensation ability, where utilizing only 1\% context samples outperforms standard ICL configurations requiring 30\%. Beyond neuroimaging, the framework’s robustness is validated by its strong generalizability across six diverse tabular datasets. Finally, PromptDx establishes a more data-efficient and clinically aligned paradigm for multimodal medical AI.

\section*{Acknowledgment}
This work was supported by the National Institute of Health (NIH) under grants R01EB022744, RF1AG077578,  R01AG064584, U19AG078109, and P30AG066530.

Data used in preparation of this article were obtained from the Alzheimer’s Disease Neuroimaging Initiative (ADNI) database (adni.loni.usc.edu). As such, the investigators within the ADNI contributed to the design and implementation of ADNI and/or provided data but did not participate in analysis or writing of this report. A complete listing of ADNI investigators can be found at: http://adni.loni.usc.edu/wp-content/uploads/how\_to\_apply/\\ADNI\_Acknowledgement\_List.pdf
%
%
%
\bibliographystyle{splncs04}
\bibliography{reference}
%




\end{document}